\pdfoutput=1
\documentclass[11pt]{article}

\usepackage[final]{EACL2023}

\usepackage{times}
\usepackage{latexsym}

\usepackage[T1]{fontenc}
\usepackage[utf8]{inputenc}

\usepackage{microtype}

\usepackage{inconsolata}

\usepackage{times}
\usepackage{latexsym}
\usepackage[T1]{fontenc}
\usepackage[utf8]{inputenc}
\usepackage{microtype}
\usepackage{verbatim}
\usepackage{multirow}

\usepackage{amsmath,amssymb,amsfonts}
\usepackage{algorithmic}
\usepackage{graphicx}
\usepackage{textcomp}

\usepackage{tabularx}
\usepackage{threeparttable}
\usepackage[]{dashrule, colortbl}
\usepackage[utf8]{inputenc} % allow utf-8 input
\usepackage[T1]{fontenc}    % use 8-bit T1 fonts
\usepackage{hyperref}       % hyperlinks
\usepackage{url}            % simple URL typesetting
\usepackage{booktabs}       % professional-quality tables
\usepackage{amsfonts}       % blackboard math symbols
\usepackage{nicefrac}       % compact symbols for 1/2, etc.
\usepackage{microtype}      % microtypography
\usepackage{amsmath,xparse, graphicx, soul}
\mathchardef\mhyphen="2D % Define a "math hyphen"
\usepackage{subfig}
\usepackage{xfakebold}
\usepackage{float}
\usepackage{todonotes}
\usepackage{colortbl}

\title{Question-Answer Sentence Graph for Joint Modeling Answer Selection}

\author{Roshni G. Iyer\textsuperscript{1}\thanks{\hspace{1em}This work was done while the author was an intern at Amazon Alexa AI.}\hspace{.3em}, Thuy Vu\textsuperscript{2}, Alessandro Moschitti\textsuperscript{2}, \and Yizhou Sun\textsuperscript{1} \\
  \textsuperscript{1}University of California, Los Angeles; Los Angeles, CA, USA \\
  \textsuperscript{2}Amazon Alexa AI; Manhattan Beach, CA, USA \\
  \texttt{\{roshnigiyer, yzsun\}@cs.ucla.edu, \{thuyvu, amosch\}@amazon.com}}
  
\begin{document}
\maketitle
\begin{abstract}
%\vspace{-.5em}
This research studies graph-based approaches for Answer Sentence Selection (AS2), an essential component for retrieval-based Question Answering (QA) systems. During offline learning, our model constructs a small-scale relevant training graph per question in an unsupervised manner, and integrates with Graph Neural Networks. Graph nodes are question sentence to answer sentence pairs. We train and integrate state-of-the-art (SOTA) models for computing scores between question-question, question-answer, and answer-answer pairs, and use thresholding on relevance scores for creating graph edges. Online inference is then performed to solve the AS2 task on unseen queries. Experiments on two well-known academic benchmarks and a real-world dataset show that our approach consistently outperforms SOTA QA baseline models.

\end{abstract}

%\vspace{-.5em}
\section{Introduction}
%\vspace{-.5em}
Automated Question Answering (QA) research has received renewed attention thanks to diffusion of Virtual Assistants. For example, Google Home, Siri, and Alexa provide general information inquiry services, while many others serve customer requests in very different application domains. Two main tasks have been widely studied: (i) Answer Sentence Selection (AS2), which, given a question and set of answer-sentence candidates, consists of selecting sentences (e.g., retrieved by a search engine) that correctly answer the question; and (ii) machine reading (MR), e.g., \cite{Chen-Fisch-2017}, which, given a question and reference text, involves finding an exact text span that answers the question.

AS2 models can more efficiently target large text databases (as they originated from the TREC-QA track \cite{voorhees99trec}) and there is evidence that they are currently used in personal assistants, e.g., Alexa \cite{DBLP:conf/sigir/MatsubaraVM20}.  

\citet{DBLP:conf/aaai/GargVM20} proposed the Transfer and Adapt (TANDA) approach, which obtained impressive improvement over SOTA for AS2, measured on two most used datasets, \mbox{WikiQA}~\cite{yang2015wikiqa} and TREC-QA~\cite{wang-etal-2007-jeopardy}. However, TANDA simply applies pointwise rerankers to individual question-answer pairs, e.g., binary classifiers, exploiting labeled out domain data (the ASNQ dataset proposed by the same authors). 

The approach above was significantly improved by the Answer Support-based Reranker (ASR/MASR) \cite{DBLP:conf/acl/ZhangVM20}, which jointly models answer candidates of each question. Essentially, the authors showed that answer candidates bring additional information for determining if a target answer $t$ is correct, and proposed an ad-hoc joint model. ASR/MASR uses TANDA as fundamental building blocks for its model, but improves on TANDA’s score via a component with AA similarity relations. Our model further improves this via a QQ similarity component integrated with a graph neural network (GNN) to capture more inter-relation dependency. 

\definecolor{officegreen}{rgb}{0.0, 0.5, 0.0}
\definecolor{mediumcarmine}{rgb}{0.69, 0.25, 0.21}
\setlength{\tabcolsep}{3pt}
\begin{table}[t]
\small
\resizebox{\linewidth}{!}{%
\begin{tabular}{|p{.4cm} p{5.8cm}|p{1.6cm}|}
\hline
$q$:&\textbf{Who won the 1967 NBA Championship ?} & \underline{$score$} \underline{$label$}\\
$c_1$:&The 1967 NBA World Championship Series was the championship series of the 1966-67 National Basketball Association season and was the conclusion of the 1967 NBA Playoffs. & \textbf{\textcolor{mediumcarmine}{0.810}} \quad \textbf{\textcolor{mediumcarmine}{0}}\vspace{.2em}\\
$c_2$: &This was the first championship series in 11 years without the Boston Celtics, who were defeated in the Division Finals by Philadelphia. & \textcolor{mediumcarmine}{0.048} \quad \textcolor{mediumcarmine}{0}\\
$c_3$: &\vspace{-.6em} The 76ers won the series over the Warriors  4-2. & \textbf{\textcolor{officegreen}{0.142}} \quad \textbf{\textcolor{officegreen}{1}}\\
\hline
\end{tabular}
}
\label{QA-input1}
\vspace{-1.5em}
\end{table}

% #############
% #############

\setlength{\tabcolsep}{3pt}
\begin{table}[t]
\scriptsize
\centering
\resizebox{\linewidth}{!}{%
\begin{tabular}{|lp{5.8cm}|lp{1.4cm}|}
\hline
$q_1$: &\textbf{Who won the 2009 Super Bowl ?} \\
$a_1$: &The Steelers defeated the Cardinals by the score of 27–23.\\
\hline
$q_2$: &\textbf{Who won Fifa World Cup 2010 ?} \\
$a_2$: &In the final, Spain, the European champions, defeated third-time finalists the Netherlands 1–0 after extra time.\\
\hline
$q_3$: &\textbf{Who won the most NBA championships ?} \\
$a_3$: &Bill Russell won 11 championships with the Boston Celtics.\\
\hline
\end{tabular}
}
\caption{An example from WikiQA, for candidates ranked by TANDA with normalized scores and labels.}
\label{QA-input2}
\vspace{-1.5em}
\end{table}

A way to go beyond ASR/MASR's approach is to consider other questions similar to the target one along with their answers for deciding over $t$.  For example, Table~\ref{QA-input2} reports a question, $q=$ \emph{Who won the 1967 NBA Championship?}, with some candidate answers, $c_1$, $c_2$, and $c_3$, sorted by a pointwise reranker. $c_1$ contains the same phrase from the question, i.e., the phrase \emph{The 1967 NBA World Championship}, which is also used by $q$ to characterize the asked information (the name of the winner). The AS2 model mistakenly ranks $c_1$ before the other candidates, as the latter do not contain the phrase above. The model answers incorrectly because for most cases matching a portion of the question in the answer is a strong evidence of its correctness. 
In contrast, the correct answer $c_3$ (which also indicates a winner) does not contain the contextual entities, i.e., NBA and 1967. 
%main entity above, thus it got a lower position in the rank. Intuitively, the model operates this choice since the majority of positive training examples have patterns \emph{similar} to $c_1$. 
 % strategy can be improved by building patterns to describe more specific cases. The main research question is how to  model more specificity. 
 
The selection ability of the AS2 model would improve if it could learn the additional context. The latter can be provided by other similar question/answer pairs. For example, in the lower part of Table~\ref{QA-input2}, we can see that the correct answer for $q_3$ (a question very similar to $q$) is $a_3$, which shows a predicate argument structure very similar to $c_3$, i.e., (\textbf{subj}:$x$, \textbf{verb}:\textit{won}, \textbf{obj}: \textit{championship}). Also $a_1$ and $a_2$ have somewhat similar structure, but most importantly all three similar questions provide evidence that it is not necessary having NBA or, in general, the name of the championship in the answer to make it correct.

%The example can provide some additional intuitions in this direction: in the lower part of the table, we note three correct question-answer pairs, where the $q_i$ are semantically similar to $q$. Interestingly, the answers, $a_1$ and $a_2$, provide evidence that winning a championship also entails to \emph{defeat/win over} another team. This information can be prioritized over the focus entity to rerank the correct answer $c_3$ to the top. Thus, an improved model should find similar examples in the training data and use them for a more effective inference.
The correctness characterization provided by the similar questions above can hardly be learned by the model from individual $(q,c)$ pairs, as the characterization does not hold in general. It can be applied relatively to the list of candidates, $\{q, c_1, c_2, c_3\}$. Therefore, these patterns should be learned from comparing  similar questions together with their list of candidates.  

In this paper, we propose to jointly model questions with their list of candidates for AS2. This enables the usage of the information from other questions, similar to the given question, together with their answers. For this purpose, we design a graph-based QA model relying on the assumption that, given question $q$ and a corresponding answer $a$, there exist other questions, $q_{i}$ and answers $a_{i}$ semantically similar to $q$, and $a$, respectively. These similar questions and answers can provide evidence to decide the correctness of $a$. To model the complex interactions among these different sentences, we use Graph Neural Networks (GNN) \cite{Gori-2005,Scarselli-2009} applied to graphs which utilize interaction among question and answer pairs.

Our main innovation regards graph construction: (i) we propose to build a different small graph for each target question, such that we improve efficiency, and effectiveness as we use specific relevant questions; (ii) as we target answer selection, which traditionally is modeled as classification of question and answer pairs, $(q,a)$, we associate graph nodes with them. We use training data to assign  pairs with positive and negative labels. (iii) We form edges between pairs using models that automatically score different types of relations: question-question (QQ), question-answer (QA), and answer-answer (AA), and then we apply thresholds to reduce the number of active edges.

We test our models over three datasets, WikiQA, TREC-QA, and WQA, where the latter is an internal dataset built with de-identified customer questions. Our GNN for QA improves the best pointwise model for AS2, i.e., TANDA, over all datasets (up to 7 absolute points on TREC-QA corresponding to 75\% of error reduction in Accuracy). It also establishes the new SOTA among joint models, improving ASR by 4 and 2 absolute points on WikiQA and TREC-QA, respectively.

%\vspace{-.5em}
\section{Related Work}
%\vspace{-0.6em}

Our work aims at improving the answer sentence selection (AS2) task in open-domain question answering (ODQA).

%\vspace{-.5em}
\paragraph{Modeling for AS2}
Previous work for AS2 modeling is typically categorized into three approaches: pointwise~\cite{shen-etal-2017-inter,DBLP:journals/corr/abs-1905-12897,DBLP:conf/aaai/GargVM20}, pairwise~\cite{conf/cikm/RaoHL16,tayyar-madabushi-etal-2018-integrating,laskar-etal-2020-contextualized}, and listwise methods~\cite{cao2007learning,conf/cikm/Bian0YCL17,DBLP:journals/corr/abs-1804-05936}. TANDA, and most other pointwise methods for AS2, however, overlook the natural existing inter-relations in the data. ASR/MASR is the current SOTA for joint modeling and considers multiple candidates for a target question. In this paper, we propose graph-based approaches for AS2, considering multiple questions and answers.

%\vspace{-.5em}
\paragraph{Graph-based Question Answering}
GNNs~\cite{Gori-2005,Scarselli-2009} have gained traction for their ability to effectively and scalably learn graph representations. Empirically, GNNs~\cite{iyer-bagnn-icdm} have achieved SOTA performance in many tasks such as node classification, link prediction, and graph classification. GNNs have been studied to improve QA in several ways. In multi-hop QA~\cite{yang-etal-2018-hotpotqa}, GNNs are used to provide a structural presentation among several entities, e.g, questions, paragraphs, sentences, named entities to facilitate reasoning~\cite{fang-etal-2020-hierarchical}. However, the link between text is always triggered by named entities or concepts.

In our work, we use entire sentence semantics to link questions or answer text. Moreover, our semantic objects are pairs and we introduce relations between them. Recently, ~\citet{DBLP:journals/corr/abs-2110-04330} has used GNNs to exploit structural relationship described in pre-built knowledge graphs to improve ranking of passages. Their approach is different from ours: they use entity knowledge graphs. More importantly, we use multiple questions, while they only use multiple passages.

For other works proposing graph-based answer selection methods, they are limited in omitting important graph dependencies such as between questions~\cite{mpge2020}, answers~\cite{WenxuanZhang-2020}, or both~\cite{WeiZhang-2021}. Further, several existing approaches require additional information to be present, such as user reputation data~\cite{Lin-2021, WeiZhang-2021}, product reviews for product-related questions~\cite{WenxuanZhang-2020}, as well as external question and answer subject knowledge~\cite{Yang-2022,Deng-2021}, assumptions that are no longer reasonable in real-world settings where this information may not be accessible. Our model overcomes these limitations, by considering the general setting where only question sentence and answer candidate information is available to effectively solve the AS2 task. 

%\vspace{-2mm}
\section{Methodology}
%\vspace{-3mm}
In this section, we first present an overview of our QA Graph Model, followed by a discussion of our graph construction procedure. Then we describe our GNN model design, and lastly detail our training and inference framework. 

%\vspace{-2mm}
\subsection{QA Graph Model Overview}
%\vspace{-2mm}
Our approach novelly reformulates the task of AS2 as a graph problem of node classification, where the goal is classify each $(q,a)$ node from test to label 1 (correct answer) or 0 (incorrect answer). To this end, our approach carefully designs a QA graph guided by TANDA-RoBERTa, which captures important inter-relations between QQ, AA, and QA sentences, discussed in Section~\ref{graph-construction}. GNNs are then used to learn final embeddings. GNN model parameters are learned during offline training, and online inference is performed for new node classification on unseen queries by turning the answer selection problem into the $(q,a)$ node classification task.

\subsection{Graph Construction}
\label{graph-construction}
Our graph construction procedure designs \textbf{E}ffective \textbf{QA} \textbf{G}raphs, EQAG. These graphs only have $(q,a)$ nodes, the construction is based on $K$-best similar questions, and there is a final step to add intra- and cross- question connections to capture inter-relation QQ, AA, and QA sentence dependencies. 

\subsubsection{EQAG node construction procedure}
Nodes are modeled as $(q,a)$ sentence pairs (for design choice details, see the Appendix), and all graph nodes are of the type $(q,a)$ (for motivation on node type choice, see Limitations section).

%\vspace{-.5em}
\subsubsection{EQAG edge construction procedure}
\begin{figure}[t]
    %\hspace{-1.5em}
    \centering
    \includegraphics[width=.9\linewidth]{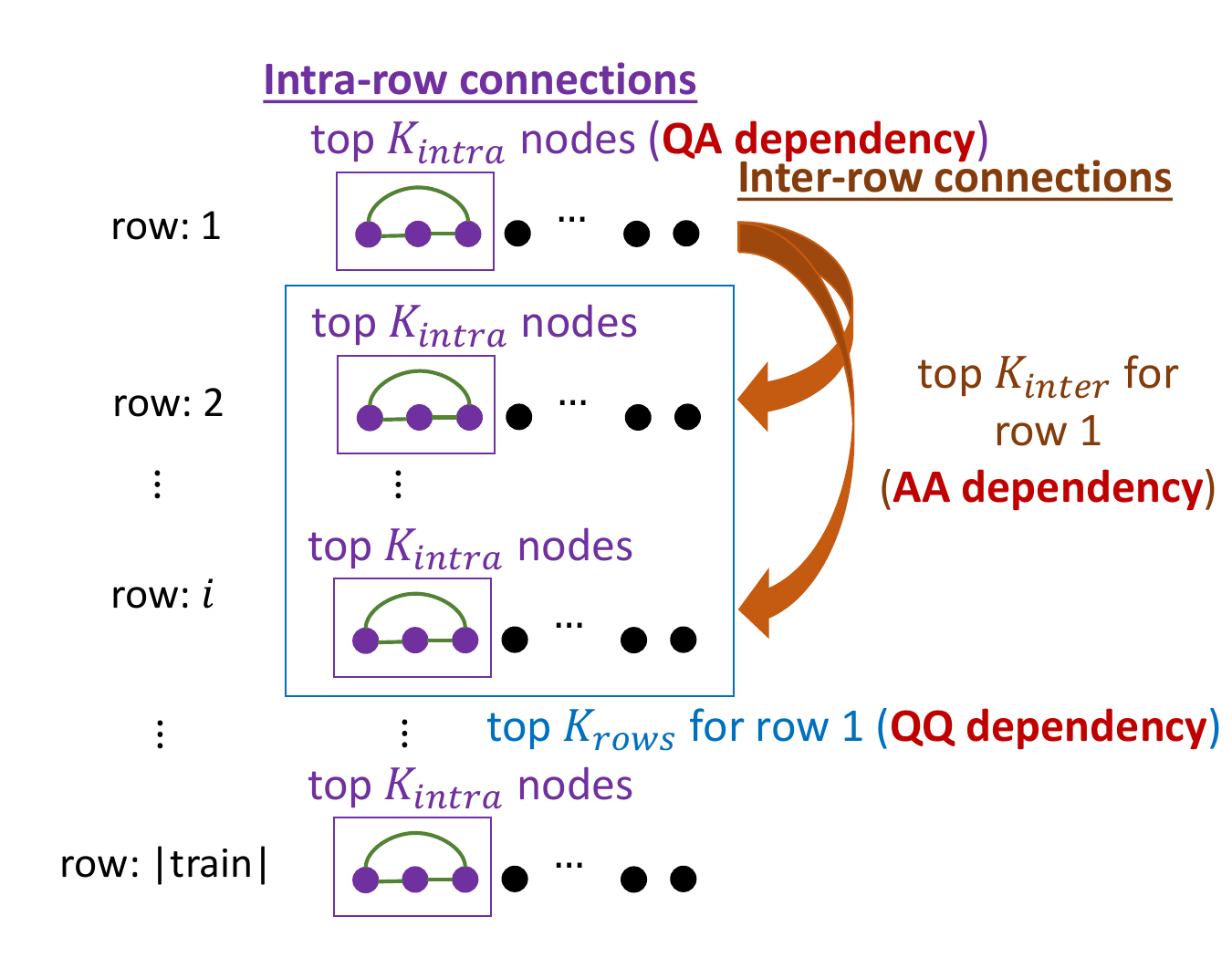}
\vspace{-.5em}
\caption{EQAG-GNN Graph Construction: each row is a question with all its answer candidates.}
    \label{fig:qa-gnn-architecture}
    \vspace{-1em}
\end{figure}

%\vspace{-.3em}
Figure~\ref{fig:qa-gnn-architecture} illustrates our approach: given query $q$, we first capture QA  dependency using the top $K_{intra}$ elements by determining which of the query's answer candidates to connect (through QA similarity). Then, we capture QQ dependency by identifying which similar questions to connect using top $K_{rows}$ (through QQ similarity). Lastly, we capture AA dependency by identifying which similar answers to connect given the nodes already have similar questions, by using the top $K_{inter}$ answers located in the top $K_{rows}$ questions (through QA similarity from one node's question to the other node's answer, which was more effective than directly using AA similarity between the nodes' answers).

\vspace{-.5em}
\paragraph{Capturing QA Dependencies.}
QA dependencies are explicitly captured via intra-row connections where for each query $q$, all corresponding query-answer nodes that have top $K_{intra}$ scores from TANDA are connected for all training data (offline learning) and test data (online learning) independently. To reduce connections between noisy data, selected top $K_{intra}$ nodes are further thresholded by $th_{intra}$. We include graph singleton nodes to enforce total question answerability coverage. 
\vspace{-.5em}
\paragraph{Capturing QQ Dependencies.}
QQ dependencies are \textit{implicitly} captured by identifying similar questions to connect using top $K_{rows}$ per query via QQ similarity. For QQ similarity, we used RoBERTa-base model trained on the open-domain Quora dataset~\footnote{\texttt{kaggle.com/c/quora-question-pairs}}. Note that use of Quora is an additional improvement in our graph construction.
\vspace{-1.5em}
\paragraph{Capturing AA Dependencies.}
AA dependencies are captured by identifying similar answers to connect using top $K_{inter}$ answers located in top $K_{rows}$ questions, via TANDA's prediction scores. Inter-row connections are further ensured to connect only training data during the offline learning stage, and performs online inference solely on unseen test data. 
Similar to intra-row connections, to reduce connections between noisy data, selected top $K_{inter}$ nodes are further thresholded by $th_{inter}$. Through empirical observation, for training nodes involved in cross-row connections, only connecting positively labeled training nodes improves model performance over connecting all labeled types of training nodes. These experiments, including other settings of inter-row connections are in Section~\ref{experiments}.
The latter also discusses hyperparameter settings for  datasets.

In practice, we find similar questions to be abundant from datasets and therefore not a limitation in finding AA dependencies. Empirically, we also find similar questions have similar answers both in terms of semantics and syntax (structure). Hypothetically, in the extreme case of having only few similar questions, EQAG-GNN model accuracy will not be compromised since AA construction additionally requires thresholding where AA similarity score must be $\geq th_{inter}$, ensuring only sufficiently similar answers have dependencies.

%\vspace{-.5em}
\subsubsection{Training Graph Construction Analysis} 
%\vspace{-.3em}
% \YS{I suggest starting the graph construction from here, and following the flow: (1) why building graph (exploit the correlation between questions and answers). (2) In the high level what are the nodes (QQ type, $(q_i,q_j)$, AA type $(a_i,a_j)$, and $(q_i,a_j)$ pairs) and relations. (3) how to create the nodes. For example, if the similarity between the two representations of the two corresponding texts are above a threshold, we will create a node for that pair. (4) how to create the links. For example, for $(q_i,q_j)$ and $(q_k,a_l)$, if the similarity between $q_i$ and $q_k$ and  $q_j$ and $a_l$ are both above some threshold. 
% }
An alternative graph construction is explicitly introducing inter-relation dependencies as nodes: $(q, q)$, $(a,a)$, and $(q,a)$. To keep graph size scalable, one may use semantically useful nodes, using QQ and AA language model similarity scores e.g., via RoBERTa with thresholding, where threshold is a hyperparameter e.g., $[0.7, 1.0]$. To form edges, one may use a similar procedure as node generation, where if any $q$ or $a$ elements between both nodes have a similarity score greater than the threshold value, to connect them. As GNNs do not update isolated nodes, one may further consider removing them to ensure scalable graph size. 

However, this above approach shows several limitations: (i) not all questions (which were supposed to be answerable) may be answered. This occurs since the constructed graph may eliminate certain queries altogether since they do not satisfy threshold score when creating nodes. Further, only connected components greater than one element are considered by the baseline. In other words, isolated nodes are removed. The edge formation process may also remove edges due to thresholding of elements between nodes, thus creating isolated nodes that are subsequently removed. The thresholding may also remove correct answer candidates, though other candidates of that query may be kept. In this case, the model will be limited to only choosing an answer candidate from existing set of candidates labeled as incorrect, leading to lower performance. 

EQAG overcomes nodes eliminated by thresholding by instead choosing top-$K$ ranked nodes for all query-answer pairs. This guarantees all queries will have prediction made by the GNN (full answerability coverage) due to their presence in the graph, and further, correct answers that may have been removed due to thresholding will be included. Moreover, having all (q, a) pairs as nodes in the graph does not negatively affect a correct node's learned representation. Initial embeddings of all (q, a) nodes are produced from TANDA-RoBERTa classifier and then updated by a GCN-GNN model to learn final embeddings which are then mapped to a binary label. While all (q, a) nodes are present, only supporting evidence (q, a) nodes are connected to eachother (which is a sparsely connected instead of a fully connected graph). Therefore, the GCN-GNN model will only update node embeddings based on sparsely available node edges. Isolated nodes are not updated via message passing (as they have no neighbor node information to aggregate from), so such a suboptimal (q, a) node's information will not affect correct nodes.

% \section{Yizhou's comments}
% \vspace{-1em}
% \YS{(1) didn't understand why it is called Initial Approach; (2) This section is about graph construction; (3) offline training and online inference is not just for graph construction.}
% \YS{a suggested writing flow: (1) overview (turn it into to a node classification problem and use GNN; offline training to learn the GNN and online inference for new node classification); (2) Talk about graph construction; (3) Talk about the GNN design including architecture and the loss; (4) talk about the training and inference. }

%\vspace{-.5em}
\section{Learning from QA graphs with GNNs}
%\vspace{-.3em}
\label{background-gnns}

GNNs broadly use message passing with graph structure learning to inform a node's representation by a recursive neighborhood aggregation scheme. A node's neighborhood aggregation considers its local context of nodes, usually set to one-hop neighbor nodes, or directly connected nodes. In this way, utilizing a node's neighborhood for learning its representation takes into effect graph connectivity, node degree, and graph features. The general framework for GNNs is as follows:  \vspace{.3em}
\begin{equation}
    \boldsymbol{h}_{i}^{(l+1)} = \sigma\big(\sum_{j \in N_{i}}f(\boldsymbol{h}_{i}^{(l)}, \boldsymbol{h}_{j}^{(l)})\big),
    \vspace{-.3em}
\end{equation}
where $\boldsymbol{h}_{i}^{(l)} \in \mathbb{R}^{d^{(l)}}$ is the feature representation of node $v_{i}$ at layer $l$ of the neural network, with dimensionality $d$. $f$ is a message-specific neural network function of incoming messages to $v_{i}$ from its neighborhood context $N_{i}$, and activation function $\sigma$, typically being $\mathrm{ReLU(\cdot)}$ for all layers but the last one which is $\mathrm{softmax(\cdot)}$. 

%\vspace{-.5em}
\paragraph{Graph Convolutional Neural Networks.} Graph Convolutional Networks (GCNs)~\cite{Kipf-2017} are a widely used class of GNNs, which have been shown to achieve superior performance on semi-supervised classification on graph-structured data. GCNs have been successfully applied to several networks including various citation network graphs, and knowledge graphs. GCN's framework is as follows where for a node $v_i = (q,a)$, where $q \in \mathrm{Q_{Train}},a \in \mathrm{A_{Train}}$, its feature $\boldsymbol{h}_{i}^{0} = \mathbf{x_i}$ is calculated by $\boldsymbol{h}_{i}^{0} = \mathrm{score_{TANDA}}(v_i)$:\\
\begin{equation}
    \boldsymbol{h}_{i}^{(l+1)} = \sigma\Big(\boldsymbol{W}_{l}^{T}\big(\sum_{j \in N_{i} \cup \{i\}} \frac{e_{j,i}}{\sqrt{m_{j}m_{i}}}\boldsymbol{h}_{j}^{(l)}\big)\Big),
    %\vspace{-.7em}
\end{equation}
where $\boldsymbol{h}_{i}^{(l)}$ are embeddings of node $v_{i}$ at layer $l \in [0, L]$, $\boldsymbol{W}_{l}$ is a layer-specific learnable weight matrix, $N_{i}$ is the set of nodes in neighborhood context of $v_{i}$, $e_{j,i}$ is edge-weight between nodes $v_{j}
\rightarrow v_{i}$, with default edge weight being $1.0$ if edge exists. $m_{i}$ and $m_{j}$ are entries of degree matrix, with $m_{i} = 1 + \sum_{j \in N_{i}} e_{j,i}$. In other words, the GNN model only uses TANDA as initial embeddings for nodes. After that, the GNN model is used to update these embeddings through multiple layers of learning, which use message passing and local neighborhoods to update the node's representation.

In this work, we explore how GNNs applied to our QA graphs are effective in learning representations of QA nodes for AS2, through their ability to inform embeddings by capturing the latent QA, QQ, and AA dependencies between nodes. 

\paragraph{GNN Loss Function.} For the task of binary node classification, GNNs use binary-cross entropy (BCE) loss for training, where only the nodes from the training set are optimized:%\vspace{-.7em}
\begin{displaymath}
    BCE = - \frac{1}{N} \sum_{i=0}^{n} y_{i} \cdot log(\hat{y_{i}}) 
    + (1 - y_{i}) \cdot log(1 - \hat{y_{i}}),
%\vspace{-.7em}
\end{displaymath}
where $y_{i}$ is the binary ground truth label for each query-answer, and $\hat{y_{i}}$ is the model's predicted probability score of the positive label, where $\hat{y_i} = \boldsymbol{h}_i^L$.

\begin{figure*}[t]
\centering
    \includegraphics[width=.77\linewidth]{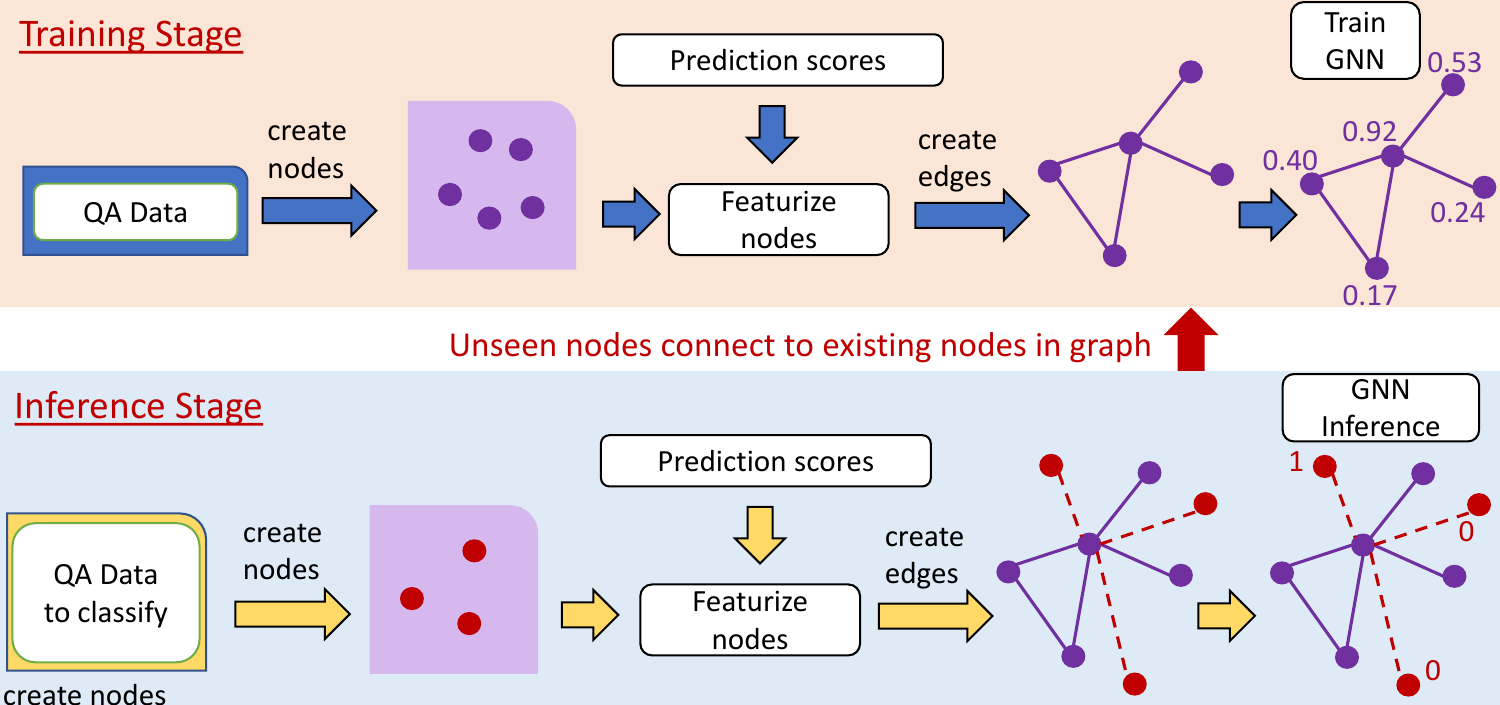}
    %\vspace{-.5em}
    \caption{EQAG-GNN Pipeline. Graph nodes are $(q,a)$ which have initial features being TANDA-RoBERTa's relevance score. During inference, for an unseen query, $(q,a)$ nodes are created and connected to existing nodes in training graph. The learned GNN is then applied to obtain classification scores and mapped to binary labels.}
    \label{fig:one-q-graph-construction}
    \vspace{-.5em}
\end{figure*}

%\vspace{-.5em}
\section{Training and Inference}

%\vspace{-2mm}
Figure~\ref{fig:one-q-graph-construction} summarizes EQAG-GNN's pipeline.

%\vspace{-.3em}
\label{subsection:training-eval}
\subsection{Training}
After the graph is constructed, described in Section~\ref{graph-construction}, it is passed to a GNN. We utilize $L=2$ layers of GCN such that initial node features are TANDA's relevance scores, representing TANDA's prediction score of how well candidate $c$ answers query $q$. After node features are updated by the GCN, the answer candidates are reranked by the learned relevance scores. Parameters of the GCN model, which include weight matrix $\mathbf{W}_{l}$, are optimized using BCE loss, described in Section~\ref{background-gnns}, such that for every query-candidate node $i$, the learned relevance score $\hat{y_{i}}$ is computed per $c$ and ranked. Further, learning rate and dropout rate are fine-tuned to the dataset. Our model's task is node classification, since each node is a query-answer to be classified with positive label (the candidate correctly answers the question) or negative label (the candidate incorrectly answers the question).

%\vspace{-3mm}
\subsection{Inference}
%\vspace{-1mm}
The inference stage is performed as online inference with only unseen query-candidate nodes. Here, each new query-answer pair forms one node, such that all possible answer candidates per query are considered. For each new query, the trained EQAG-GNN model leverages its neighborhood information of certain existing $(q,a)$ nodes from training that it is connected to, based on top $K_{inter}$ rows in EQAG. As computation of the top $K_{inter}$ rows uses QQ, AA, and QA semantic similarity, the GNN model leverages these important inter-relation dependencies when classifying a new query-candidate node. Further, as the model's offline training is strictly performed on training nodes with no additional data leveraged, it is ensured that there is no information leak. Once EQAG-GNN predicts scores of unseen nodes, scores are ranked to find the candidate with the highest learned relevance score. Then, the soft-valued score is converted to a binary valued label where only the highest scoring candidate is assigned label 1 and the remaining candidates are label 0.   

For effectiveness, during inference, we also do not fully connect all test answer candidates of a query when building the inference graph but instead use ranked top-$K$ candidates guided by TANDA-RoBERTa's prediction score to form edge connections. This is suggested by our ablation study: more noise is included in the model when the QA graph becomes too densely connected. At the same time, an overly sparsely connected graph will also lead to less accuracy during inference of the model since GNN's message passing component will not be as effective. Therefore, we choose $K$ as tunable hyperparameter on the target dataset.

%\vspace{-.5em}
\paragraph{GNN parameter size:} Our GNN is also efficient in learned parameter size, as it uses the same parameter size as the efficient TANDA model and its GNN model parameter size is also minimal. Specifically, for the GCN GNN model, we used parameter size of $d \cdot d + n \cdot d$, where $d$ is embedding size, and $n$ is number of graph nodes. In our model setting, $d=1$ and $n$ is dataset specific, which in practice is in magnitude of a few thousand nodes in total, or in magnitude of a few hundred nodes per question on average. Specifically, $n$ is around 26K, 52K, and 426K in total for the datasets of WikiQA, TREC-QA, and WQA respectively, which on average per question is 31, 42, and 121 nodes for WikiQA, TREC-QA, and WQA respectively. 

\vspace{-1em}
\section{Experiments}
%\vspace{-.3em}
\label{experiments}

Here, we compare EQAG-GNN against SOTA QA models. Then, we show an ablation study for EQAG-GNN for the best hyperparameters. Finally, we discuss a case study with error analysis.

\begin{table}
\center
\small{
\begin{tabular}{|c|c|c|c|c|}
% \hline
%  \multicolumn{5}{|c|}{Country List}  \\
%  \hline
\hline
 \multicolumn{4}{|c|}{\textbf{WikiQA}}\\
 \hline
 Statistics & Train & Dev & Test \\
 \hline
 \#Q & 873 & 121 & 237 \\
 \hline
  \#A+ & 1,040 & 140 & 293 \\
 \hline
  \#A- & 7,632 & 990 & 2,058 \\
 \hline
 \multicolumn{4}{|c|}{\textbf{TREC-QA}}\\
 \hline
 Statistics & Train & Dev & Test \\
 \hline
 \#Q & 1,227 & 65 & 68 \\
 \hline
  \#A+ & 6,388 & 205 & 248 \\
 \hline
  \#A- & 46,974 & 912 & 1,194 \\
 \hline
 \multicolumn{4}{|c|}{\textbf{WQA}}\\
 \hline
 Statistics & Train & Dev & Test \\
 \hline
 \#Q & 3,519 & 648 & 717 \\
 \hline
  \#A+ & 42,739 & 6,147 & 6,356 \\
 \hline
  \#A- & 96,049 & 10,034 & 11,539 \\
 \hline
\end{tabular}}
\vspace{.5em}
\caption{\label{table:dataset-properties}AS2 dataset statistics, number questions, positive/negative answers, from official train/dev/test splits.}
\vspace{-.5em}
\end{table}

%\vspace{-.5em}
\subsection{Datasets}
%\vspace{-.3em}

Table~\ref{table:dataset-properties} shows a description of the datasets. 

%\vspace{-.5em}
\paragraph{WikiQA:} \hspace{-1em} WikiQA~\citep{yang2015wikiqa} is an AS2 dataset containing data with form label-question-answer such that labels are binary, indicating existence of positive or negative QA pair, and there are several answer candidates for a question. Data comes from Bing query logs over Wikipedia where answers are manually labeled. We follow the most used setting: training with all questions having at least one correct answer, and validating and testing with all questions with at least one correct and one incorrect answer.

%\vspace{-.5em}
\paragraph{TREC-QA:} \hspace{-1em} TREC-QA~\cite{wang-etal-2007-jeopardy} is an AS2 dataset containing data in a format like WikiQA of label-question-answer. We use the same splits of the original data, following the setting of previous work~\citep{DBLP:conf/aaai/GargVM20}.

%\vspace{-.5em}
\paragraph{WQA:} \hspace{-1em} WQA~\citep{Zhang-wdrass-2022} is an AS2 dataset built with anonymized customers' utterances from a popular personal assistant. The dataset was built as part of the effort to improve understanding and benchmarking in ODQA. The creation process includes steps: (i) given a set of questions collected from the web, a search engine is used to retrieve up to 1,000 web pages from an index containing millions of pages. (ii) From retrieved documents, all candidate sentences are extracted and ranked using AS2 models. Finally, (iii) top candidates for each question are manually assessed as correct or incorrect by human judges. This allowed obtaining higher average number of correct answers with richer variety from multiple sources, shown in Table~\ref{table:dataset-properties}. Data is in a format similar to WikiQA of label-question-answer. For consistency with standard QA datasets, we filter out WQA for all non-answerable questions, or questions with only negative answer candidate choices.

%\vspace{-.5em}
\subsection{Setup}
%\vspace{-.3em}

%\vspace{-.2em}
\paragraph{Metrics:} 
Performance of QA systems is typically measured with Accuracy being percentage of correct responses. This is also referred to as Precision-at-1 (P@1) in the context of reranking, while standard Precision and Recall are not meaningful as the system does not abstain from providing answers.

%\vspace{-.5em}
\paragraph{Implementation details:}
\label{implementation-details} 
As the basic language model for our systems, we used the TANDA checkpoint, which is the SOTA AS2~\citep{DBLP:conf/aaai/GargVM20}. This is a pre-trained RoBERTa-base, further fine-tuned on ASNQ data~\footnote{Available at~\texttt{github.com/alexa/wqa\_tanda}}. We use the same reported optimal hyperparameter settings~\citep{DBLP:conf/aaai/GargVM20}. Specifically, 4 Tesla V100 GPUs with 32GB for training and evaluation batch sizes of 32, with the maximum sequence length 128, and learning rate of 1e-6 for adapt step on the target dataset. We adopt Adam optimizer~\citep{Kingma-2015} with learning rate of 2e-5 for the transfer step on ASNQ. 

%\vspace{-.5em}
\paragraph{Model Parameters: } To construct our QA graph, we used 8 Tesla V100 GPUs with 32GB with training batch size of 256. We utilized TANDA's configuration to guide the initial graph features, as described in Section~\ref{implementation-details}. We then utilize the GCN model to complete the final step of model training, such that the node's embedding dimension size is set to 1, initially being TANDA's prediction scores. Learning rates were hyperparameters tuned from {1e-6, 2e-6, 5e-6, 1e-3, 2e-5}, and used for WikiQA/TREC-QA/WQA with 1e-3, 1e-3, and 1e-6. Number of layers were hyperparameters tuned from {2, 4, 8, 16} and we utilized 2 layers for datasets. Training and eval batch sizes were 32.

\subsection{Experiments with EQAG}
%\vspace{-.2em}
We consider four EQAG-GNN model variants:

 \textbf{RI} (Random Initialization): node features are randomly initialized with Gaussian uniformly random distribution between $[0.0, 1.0]$. This is to test impact that TANDA's model has on guiding EQAG-GNN's learned embeddings. For all variant models, we use hyperparameter settings from Section~\ref{hyperparameter-settings}. 
    
\textbf{TA} (Train All): both positively and negatively labeled training examples are considered to connect nodes between rows (inter-row connections).
    
\textbf{T$+$ }(Train on positives): only the positively labeled training examples are considered to form the connections of nodes between rows. This approach seems to reduce the noise that incorrect training data may introduce to the model, when learning embeddings for all training data considered together. 
    
\textbf{T$\pm$ }(Train positive and negative individually):  for inter-row connections, both positively and negatively labeled training examples are used for connections, but considered separately. Specifically, we propagate both positive and negative node information throughout the network as different subcomponents. Positively labeled training data connects to the top $K_{intra}$ connected test nodes, while negatively labeled training data connects to the bottom $K_{intra}$ test nodes, which are isolated while ensuring that node features are less than $th_{intra}$. This may guide the model to learn better embeddings, as negative nodes will be influenced more by their negative neighbor context, while positive nodes will be influenced more by their positive neighbor context during the GNN message passing stage. This helps better distinguish embeddings between positive and negative test nodes during inference as overlap between local contexts of positive and negative nodes will be minimized.   

% \begin{table}[t]
% \center
% \small{
% \begin{tabular}{|c|c|c|c|c|}
% % \hline
% %  \multicolumn{5}{|c|}{Country List}  \\
% %  \hline
% \hline
%  \multicolumn{5}{|c|}{Sample 1}\\
%  \hline
%  Metric & P@1 & \#Q Inc. & \#Q Answerable 
%  & \#Q\\
%  \hline
% TANDA & 0.8792 & 18 & 149 considered & 237 \\ 
%  \hline
% GQAG & \textbf{0.8926} & 16 & 149 & 237 \\ 
%  \hline
% \end{tabular}
% \vspace{-.5em}
% \caption{\label{table:exp1}Comparison between TANDA and GNN applied to GQAG on a sample from the WikiQA test set}}
% \vspace{-.5em}
% \end{table}

% \begin{table}
% \center
% \small{
% \begin{tabular}{|c|c|c|c|c|}
% % \hline
% %  \multicolumn{5}{|c|}{Country List}  \\
% %  \hline
% \hline
%  \multicolumn{5}{|c|}{Sample 2}\\
%  \hline
%  Metric & P@1 & \#Q Inc. & \#Q Answerable 
%  & \#Q\\
%  \hline
%  TANDA & 0.8863 & 15 & 132 considered & 237 \\ 
%  \hline
%  GQAG & \textbf{0.9091} & 12 & 132 & 237 \\ 
%  \hline
% \end{tabular}
% \vspace{-.5em}
% \caption{Comparison between TANDA and GNN applied to GQAG on another sample}
% \label{table:exp2}
% }
% \vspace{-1em}
% \end{table}
% \small{TANDA (pre-trained model)}

%\vspace{-.5em}
\paragraph{Choosing EQAG-GNN hyperparameters}
\label{hyperparameter-settings}

\begin{figure*}%
    % \centering
    \subfloat[\centering top $K_{intra}$ and top $K_{inter}$ using WikiQA]{{\includegraphics[height=5cm, width=7.5cm]{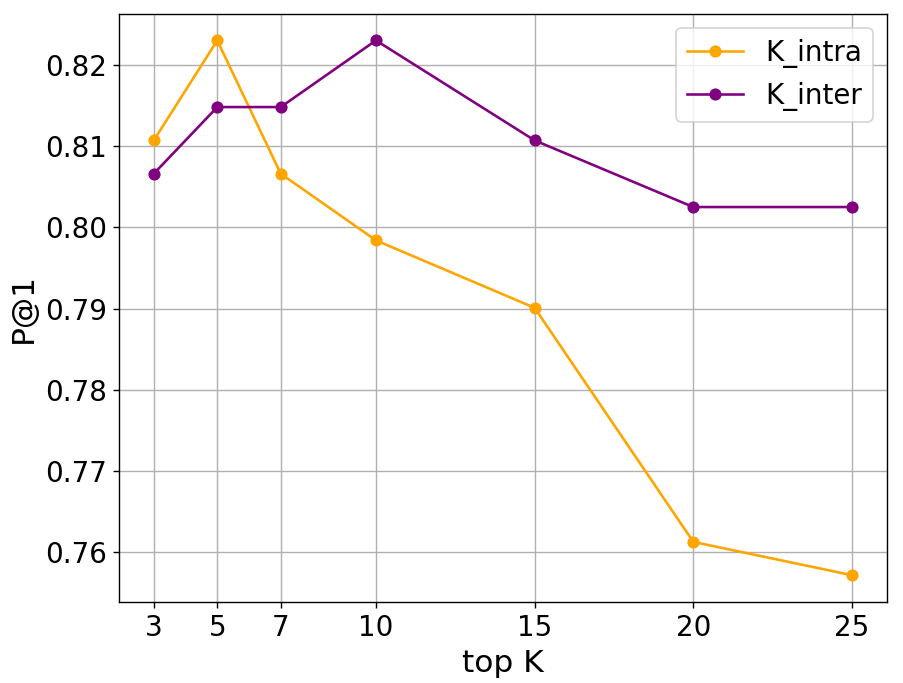} }}\label{fig:ablation-K}%
    \hspace{1em} 
    \subfloat[\centering $th_{intra}$ and $th_{inter}$ using WikiQA] { {\includegraphics[height=5cm,width=7.5cm]{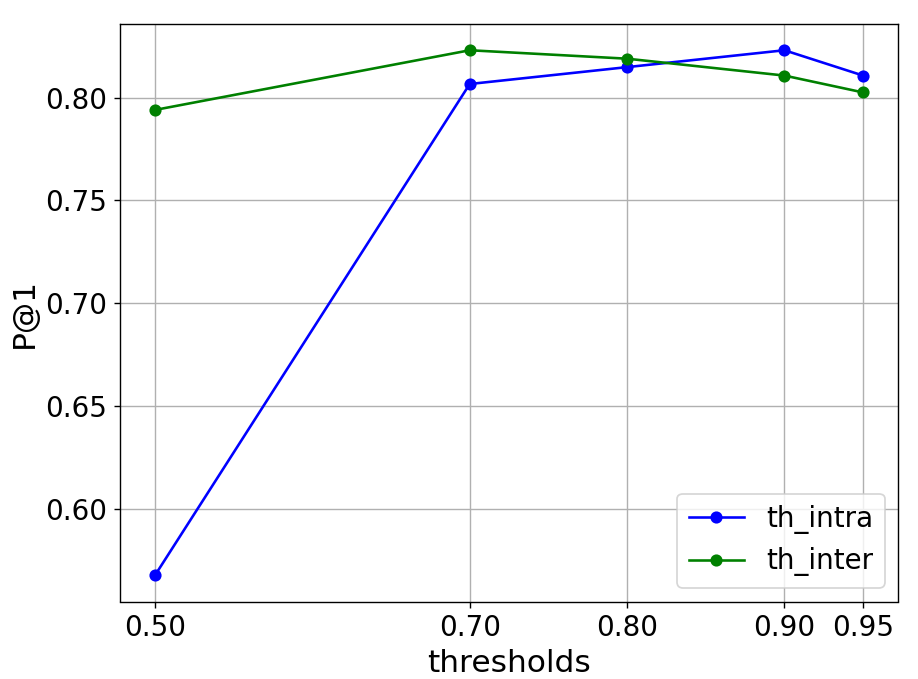} }}\label{fig:ablation-thresholds}%
    \caption{EQAG-GNN Hyperparameters on WikiQA}%
    \vspace{-1em}
    \label{fig:EQAG-hyperparameters}
\end{figure*}

% \begin{figure}[t]
%     \centering
%     \includegraphics[width=0.95\linewidth,height=0.65\linewidth]{fig/ablation-K.PNG}
%     \vspace{-.5em}
% \caption{top $K_{intra}$ and top $K_{inter}$ using WikiQA}
%     \label{fig:ablation-K}
% \vspace{-.5em}
% \end{figure}
% \begin{figure}[t]
%     \centering
%     \includegraphics[width=0.95\linewidth,height=0.65\linewidth]{fig/ablation-thresholds.PNG}
%     \vspace{-.5em}
% \caption{$th_{intra}$ and $th_{inter}$ using WikiQA}
%     \label{fig:ablation-thresholds}
%     \vspace{-1em}
% \end{figure}

We run ablation studies for hyperparameter tuning on the WikiQA dataset, evaluated on the validation set. For each hyperparameter tuned, we fix the values of all other hyperparameters to their optimal setting. The optimal settings of the variables, as shown in Figures~\ref{fig:EQAG-hyperparameters}(a) and~\ref{fig:EQAG-hyperparameters}(b), are $K_{intra} = 5$, $K_{inter} = 10$, $th_{intra} = 0.70$, and $th_{inter} = 0.90$. It can be observed that $K_{intra}$ and $th_{intra}$ may affect the P@1 accuracy more than $K_{inter}$ and $th_{inter}$, perhaps because they directly impact how the intra-row connections between answer candidates of test queries are formed, which greatly influences the learned embeddings of each test node. 
\vspace{-2em}
\paragraph{Comparative Results} 
Table~\ref{table:eval_1} reports P@1 accuracies of different SOTA QA models for ODQA on AS2 evaluated on WikiQA, TREC-QA, and WQA datasets. Models include TANDA, ASR, MASR, KGAT, EQAG-GNN model variants, described above, the SOTA GNN-based model for AS2, BR-MPGE-AS~\cite{mpge2020}, the CNN-based sentence similarity model, L.D.C.~\cite{ldc}, the Bi-LSTM CNN-based model which explicitly models pairwise word interactions, PWIM~\cite{pwim}, the hyperbolic space embedding model, HyperQA~\cite{hyperqa}, and the CNN-based latent clustering (LC) language model (LM), Comp-Clip+LM+LC~\cite{DBLP:journals/corr/abs-1905-12897}. All models use RoBERTa-base pre-trained checkpoint. The table shows QA-GNN consistently achieves highest performance among all models and all datasets on P@1 metric. For example, it outperforms TANDA by around 4, 7, and 3 absolute percent points in P@1 on WikiQA, TREC-QA, and WQA, respectively. Appendix tables~\ref{table:eval_2} and~\ref{table:eval_3} further report Maximum a Posteriori (MAP) and Mean Reciprocal Rank (MRR) scores, showing EQAG-GNN also achieves similar performance gains.   

% \vspace{-2mm}

Results show 100\% coverage of answerable questions and boosted accuracy of the approach. The ranking mechanism also improves graph connectivity, since all top-ranked question-answer pairs are included, though some nodes may be isolated. This technique greatly improves the amount of information needed by the GNN, compared to when only connected components without isolated nodes were considered. Finally, since the approach needs to start from weights learned by TANDA as random initialization, GNN-RI, produces very low results. Experiment results confirm effectiveness of modeling sentence-level semantics via graph-based models for AS2 (our approach) through comprehensive comparison of SOTA methods on AS2 including graph-based AS2 methods that model entity-level semantics. Further, our model consistently achieves significant improved performance against all baselines on all comprehensive metrics for AS2 (P@1, MAP, MRR), showing both accuracy and quality of QA pair ranking learned. 

% Regarding the efficiency, the major burden is the construction of the graph for each test question, as the GNN processing is typically very fast. Building the graph requires roughly 5 seconds for each question, which can be prohibitive for online applications. However, the main complexity is due to the retrieval of the top $K$-questions for each target query, which we perform with a RoBERTa cross-encoder. Methods like DPR~\cite{karpukhin-etal-2020-dense} can highly speed up the approach and make it comparable to the document retrieval step, which is part of any retrieval-based QA system. Finally, the approach as is, can be effectively applied to operate offline processing such those required by community QA, forums, and so on.

\begin{table}[t]
\center
\small{
\begin{tabular}{|c|c|c|c|}
\hline
Model	 & \textbf{WikiQA} & \textbf{TREC-QA} & \textbf{WQA} \\
\hline
  & \multicolumn{3}{c|}{P@1} \\
 \hline
%  Reranker by Garg et al., 2020 & - & - \\
%  \hline
BR-MPGE-AS & 0.835 & 0.912 & 0.600 \\
\hline
L.D.C. Model & 0.549 & 0.618 & 0.402 \\
\hline
PWIM & 0.823 & 0.824 & 0.582 \\
\hline
HyperQA & 0.827 & 0.853 & 0.598 \\
\hline
Comp-Clip + LM + LC & 0.827 & 0.838 & 0.590 \\
\hline
 TANDA & 0.823 & 0.912 & 0.651 \\
 \hline
  KGAT ($k=2$) & 0.844 & 0.941 & -- \\
 \hline
  ASR ($k=3$) & 0.844 & 0.971 & -- \\
 \hline
  MASR ($k=3$) & 0.823 & 0.927 & -- \\
 \hline
EQAG-GNN (RI) & 0.309 & 0.412 & 0.223 \\
 \hline
%   QA-GNN [TANDA Reranker only] & - & - \\
%  \hline
 EQAG-GNN (TA) & 0.840 & 0.956 & 0.671\\
 \hline
 EQAG-GNN (T$+$) & 0.860 & \textbf{0.985} & 0.676 \\
 \hline
 EQAG-GNN (T$\pm$) & \textbf{0.864} & \textbf{0.985} & \textbf{0.679}\\
 \hline
\end{tabular}
% \vspace{.5em}
\caption{\label{table:eval_1} P@1 evaluation of GNN applied to EQAG and leading baselines. The best results are bold-faced.}}
\vspace{-1.0em}
\end{table}

% ##############
% ##############
\vspace{-1mm}
\section{Case Study}
\vspace{-2mm}
We provide a case study and error analysis comparing performance of EQAG with TANDA. Table~\ref{table:case-study1} reports question, $q=$ \emph{How is Jameson Irish Whiskey made?}, with candidate answers, $c_1$, through $c_7$, for which the AS2 model has to pick out the best candidate answer. Table~\ref{table:case-study2} reports prediction scores learned by EQAG, and TANDA, per candidate answer and ground truth label.

\setlength{\tabcolsep}{3pt}
\begin{table}[t]
\tiny
\resizebox{\linewidth}{!}{%
\begin{tabular}{|p{.4cm} p{5.8cm}|p{1.6cm}|}
\hline
$q$:&\textbf{How is Jameson Irish Whiskey made ?}\\
$c_1$:&Jameson is similar in its adherence to the single distillery principle to the single malt tradition  but Jameson blends column still spirit with Single pot still whiskey  a combination of malted barley with unmalted or `` green '' barley distilled in a pot still.\\
$c_2$: &Jameson is a single distillery Irish whiskey produced by a division of the French distiller Pernod Ricard.\\
$c_3$: &The company was established in 1780 when John Jameson established the Bow Street Distillery in Dublin.\\
$c_4$: &Originally one of the six main Dublin Whiskeys  Jameson is now distilled in Cork  although vatting still takes place in Dublin.\\
$c_5$: &With annual sales of over 31 million bottles  Jameson is by far the best selling Irish whiskey in the world  as it has been sold internationally since the early 19th century when John Jameson along with his son (also named John) was producing more than a million gallons annually.\\
$c_6$: &Portraits of John and Margaret Jameson by Sir Henry Raeburn are in the collection of the National Gallery of Ireland.\\
$c_7$: &Jameson was Scottish  a lawyer from Alloa who had married Margaret Haig  a sister of the brothers who founded the main Haig firms and related to the Steins  a Scottish distilling family with interests in Dublin.\\
\hline
\end{tabular}
}
%\vspace{-0.5em}
\caption{\label{table:case-study1} Case study example of $(q, a)$ from WikiQA.}
% \vspace{-1em}
\end{table}

% ##############
% ##############

\setlength{\tabcolsep}{3pt}
\begin{table}[t]
\small
\resizebox{\linewidth}{!}{%
\begin{tabular}{|p{0.5cm} p{2.5cm} p{2.5cm}p{1.6cm}|}
\hline
& \underline{$EQAG \ score$} & \underline{$TANDA \ score$} &  \underline{$label$}\\
$c_1$:& \quad{\textcolor{officegreen}{\textbf{\underline{0.192}}}} & \quad{\textcolor{officegreen}{\textbf{0.108}}} & \quad{\textcolor{officegreen}{\textbf{1}}}\\
$c_2$:& \quad{\textcolor{mediumcarmine}{\textbf{0.158}}} & \quad{\textcolor{mediumcarmine}{\textbf{\textbf{\underline{0.505}}}}} & \quad{\textcolor{mediumcarmine}{\textbf{0}}}\\
$c_3$:& \quad{0.103} & \quad{0.110} & \quad{0}\\
$c_4$:& \quad{0.130} & \quad{0.266} & \quad{0}\\
$c_5$:& \quad{0.140} & \quad{0.009} & \quad{0}\\
$c_6$:& \quad{0.128} & \quad{$\sim$ 0} & \quad{0}\\
$c_7$:& \quad{0.148} & \quad{$\sim$ 0} & \quad{0}\\
\hline
\end{tabular}
}
% \vspace{0.5em}
\caption{\label{table:case-study2} EQAG, TANDA scores for Table~\ref{table:case-study1}, with normalized predicted candidate answer scores underlined.}
\vspace{-1em}
\end{table}

%\vspace{-3mm}
\subsection{Error Analysis}
%\vspace{-1mm}
TANDA mistakenly ranks $c_2$ of label $0$ before other candidates, while EQAG correctly ranks $c_1$ of label $1$ above other candidates. Though phrase \textit{Jameson Irish Whiskey} is important to the question, semantic intent of the question is how it is \textit{made}. While TANDA recognizes importance of phrase \textit{Jameson Irish Whiskey}, it does not learn necessary context of what the question asks for. As such, it picks the candidate describing what the whiskey is rather than how it is made. EQAG, however, places attention on both \textit{Jameson Irish Whiskey} and question context \textit{made}, while also learning that entire phrase \textit{Jameson Irish Whiskey} may not need to be present in the candidate sentence as long as there is some indication of referring to the item e.g., \textit{Jameson}. 

As shown in Table~\ref{table:case-study3}, EQAG effectively learns from similar questions and its corresponding answer candidates that it sees during training to recognize important semantic characteristics of the question. For example, all three supporting questions e.g., $q_1$ through $q_3$ are about how various items (single malt scotch, bourboun, root beer) are made. Further, correct corresponding answers contain information about \textit{both} item string name as well as how it is made. Regarding item string name, it is not necessary for the entire string name to be present as long as appropriate substring referring to this item is there. For example, $a_1$ refers to \textit{single malt scotch} as simply \textit{Scotch}, which EQAG also learns in order to identify that \textit{Jameson Irish Whiskey} may be referred to as \textit{Jameson}.

% ##############
% ##############

\setlength{\tabcolsep}{3pt}
\begin{table}[t]
\small
\resizebox{\linewidth}{!}{%
\begin{tabular}{|p{.4cm} p{5.8cm}|p{1.6cm}|}
\hline
 & & ($q$, $q_i$) sim \\
 \hline
$q_1$:&\textbf{How is single malt scotch made ?} & \quad{0.743}\\
$a_1$:&As with any Scotch whisky, a single malt Scotch must be distilled in Scotland and matured in oak casks in Scotland for at least three years (most single malts are matured longer). & \\
\hline
$q_2$:&\textbf{What is bourboun made of ?} & \quad{0.622}\\
$a_2$:& Bourbon whiskey is a type of American whiskey– a barrel-aged distilled spirit made primarily from corn. & \\
\hline
$q_3$:&\textbf{How is root beer made ?} & \quad{0.498}\\
$a_3$:& Root beer is a carbonated, sweetened beverage, originally made using the root of a sassafras plant (or the bark of a sassafras tree) as the primary flavor. & \\
\hline
\end{tabular}
}
%\vspace{0.5em}
\caption{\label{table:case-study3} EQAG learned similar questions to question from Table~\ref{table:case-study1}, with corresponding absolute similarity scores between $(q, q_i)$, $i \in [1,3]$}
\vspace{-1.5em}
\end{table}

% \vspace{-.5em}
\section{Conclusions}
\vspace{-1.5mm}
To our knowledge, our model is the first graph-based approach for jointly modeling sentence-level semantics of question-answer pairs for AS2 as an offline processing application, such as those required by community QA, forums, etc. This is different from previous methods using graphs, e.g., MultiHop or Graph-based QA, which mainly model semantics via entities. Our approach builds query-specific small-scale training graphs for offline learning, through $(q,a)$ pairs as nodes, and edges encoding relations between members of pairs to capture both supporting question-question, and answer-answer dependencies. Further, we demonstrate that our approach achieves significant performance gains over existing SOTA models on AS2 for metrics of P@1, MAP, and MRR.

\section*{Acknowledgments}
\vspace{-1mm}
The work of the first and last authors was partially supported by NSF 2211557, NSF 1937599, NSF 2119643, NASA, Okawa Foundation Grant, Amazon Research Awards, Cisco research grant, Picsart Gifts, and Snapchat Gifts.

\section*{Limitations}

Our proposed model is efficient as its complexity is comparable to SOTA retrieval models like TANDA-RoBERTa. However, we note that out of our model components, the main complexity is from graph construction for the offline learning stage when constructing query-specific small-scale training graphs as opposed to graph processing, since GNN processing is typically fast. This is due to retrieval of top $K$-questions for each target query, which we perform with a RoBERTa cross-encoder. While the graph building takes roughly a few seconds per question in practice, given the scope of our problem of investigation for \textit{offline} training and \textit{online} learning, our model is efficient. 

Further, while we capture Q-Q and A-A dependencies to form edge connections between the $(q,a)$ nodes, the fact that the nodes in EQAG are all of the form $(q,a)$ may be seen as a limitation. However, in the context of QA, the model deciding if an answer is correct or not for a question is trained over $(q,a)$ pairs. This means that most information is captured by the pair, which is seen as the whole object. In general, we aim at modeling the similarity between pairs in the graph as we want to learn the patterns that make a pair correct. A graph having nodes as pairs directly enables this kind of learning, with also the great advantage that cross encoding two pieces of text in a transformer always produces a much higher accuracy than separated encoding of question and answer. 

As future work, we plan to investigate building a model for learning graph topologies~\cite{iyer-dgs-kdd}, and other online processing applications e.g., document retrieval, by exploring methods like DPR~\cite{karpukhin-etal-2020-dense} to further speed up offline graph construction.

% Entries for the entire Anthology, followed by custom entries
% \bibliography{anthology,custom}
\vspace{-2mm}
\bibliographystyle{acl_natbib}
\bibliography{acl2022,all-asr}

\appendix

\begin{table}[h]
\center
\small{
\begin{tabular}{|c|c|c|c|}
\hline
Model	 & \textbf{WikiQA} & \textbf{TREC-QA} & \textbf{WQA} \\
\hline
  & \multicolumn{3}{c|}{MAP} \\
 \hline
%  Reranker by Garg et al., 2020 & - & - \\
%  \hline
BR-MPGE-AS & 0.867 & 0.897 & 0.661 \\
\hline
L.D.C. Model & 0.706 & 0.771 & 0.582 \\
\hline
PWIM & 0.709 & 0.758 & 0.550 \\
\hline
HyperQA & 0.712 & 0.784 & 0.561 \\
\hline
Comp-Clip + LM + LC & 0.764 & 0.868 & 0.610 \\
\hline
 TANDA & 0.889 & 0.914 & 0.653 \\
 \hline
  KGAT ($k=2$) & 0.899 & 0.916 & -- \\
 \hline
  ASR ($k=3$) & \textbf{0.901} & 0.928 & -- \\
 \hline
  MASR ($k=3$) & 0.889 & 0.920 & -- \\
 \hline
EQAG-GNN (RI) & 0.384 & 0.485 & 0.301 \\
 \hline
%   QA-GNN [TANDA Reranker only] & - & - \\
%  \hline
EQAG-GNN (TA) & 0.869 & 0.897 & 0.656\\
 \hline
 EQAG-GNN (T$+$) & \textbf{0.901} & 0.926 & 0.658 \\
 \hline
 EQAG-GNN (T$\pm$) & \textbf{0.901} & \textbf{0.941} & \textbf{0.662}\\
 \hline
\end{tabular}
\vspace{-.5em}
\caption{\label{table:eval_2}MAP evaluation of GNN applied to EQAG and leading baselines. The best results are bold-faced.}}
\vspace{-1em}
\end{table}

\section{Additional Experiment Results}
Table~\ref{table:eval_2} reports MAP scores and Table~\ref{table:eval_3} reports MRR scores for EQAG variant models, as well as SOTA QA models for AS2. The tables further show that QA-GNN consistently achieves the highest performance among all models and all
datasets on MAP and MRR metrics. For example, on MAP, it outperforms TANDA by around 1, 3, and 1 absolute percent points on WikiQA, TREC-QA, and WQA, respectively. On MRR, it outperforms TANDA by around 2, 3, and nearly 2 absolute percent points on WikiQA, TREC-QA, and WQA, respectively.

% #################
% #################

\begin{table}[t]
\center
\small{
\begin{tabular}{|c|c|c|c|}
\hline
Model	 & \textbf{WikiQA} & \textbf{TREC-QA} & \textbf{WQA} \\
\hline
  & \multicolumn{3}{c|}{MRR} \\
 \hline
%  Reranker by Garg et al., 2020 & - & - \\
%  \hline
BR-MPGE-AS & 0.879 & 0.912 & 0.669 \\
\hline
L.D.C. Model & 0.723 & 0.845 & 0.598 \\
\hline
PWIM & 0.723 & 0.822 & 0.593 \\
\hline
HyperQA & 0.727 & 0.865 & 0.630 \\
\hline
Comp-Clip + LM + LC & 0.784 & 0.928 & 0.636 \\
\hline
 TANDA & 0.901 & 0.952 & 0.681 \\
 \hline
  KGAT ($k=2$) & 0.912 & 0.965 & -- \\
 \hline
  ASR ($k=3$) & 0.912 & 0.982 & -- \\
 \hline
  MASR ($k=3$) & 0.902 & 0.963 & -- \\
 \hline
EQAG-GNN (RI) & 0.359 & 0.544 & 0.371 \\
 \hline
%   QA-GNN [TANDA Reranker only] & - & - \\
%  \hline
 EQAG-GNN (TA) & 0.907 & 0.956 & 0.690\\
 \hline
 EQAG-GNN (T$+$) & 0.916 & 0.971 & 0.697 \\
 \hline
 EQAG-GNN (T$\pm$) & \textbf{0.924} & \textbf{0.983} & \textbf{0.699}\\
 \hline
\end{tabular}
\vspace{-.5em}
\caption{\label{table:eval_3}MRR evaluation of GNN applied to EQAG and leading baselines. The best results are bold-faced.}}
\vspace{-1em}
\end{table}

\section{Nodes Modeled as Sentence Pairs}
In the context of QA, we decide if an answer is correct or not for a question by training QA classifiers. This means that most information is captured by the pair, which is seen as the whole object. In general, we aim at modeling the similarity between pairs in the graph as we want to learn the patterns that make a pair correct. A graph having nodes as pairs directly enable this kind of learning, with also the great advantage that cross encoding two pieces of text in a transformer always produces a much higher accuracy than separated encoding of question and answer. 

\end{document}